\documentclass[twocolumn,11pt,cleanfoot, cleanhead, final]{asme2e}

\usepackage{color}
\usepackage{booktabs}
\usepackage{graphicx}
\usepackage{dblfloatfix}
\usepackage{multirow} 
 \usepackage{array}
 \usepackage{float}
 \usepackage{longtable}
 \usepackage{balance}
 \usepackage{array}
\usepackage{booktabs}
 
 \usepackage[table,xcdraw]{xcolor}
\usepackage{appendix}
\usepackage{adjustbox}
\usepackage{amsmath}
\usepackage{flafter} 
\usepackage{multirow}
\usepackage{rotating}
\usepackage{amsmath}
\usepackage{amssymb}
\usepackage{subfigure}
\usepackage{hyperref}
\usepackage[ruled,norelsize]{algorithm2e}
\usepackage[compact]{titlesec}
\titlespacing{\section}{0pt}{5ex}{0.5ex}
\titlespacing{\subsection}{0pt}{3ex}{1ex}
\titlespacing{\subsubsection}{0pt}{1ex}{2ex}
\usepackage{multirow}
\usepackage[table,xcdraw]{xcolor}
\usepackage{setspace}

\title{On the effectiveness of Large Language Models \\ in the mechanical design domain}

\author{Daniele Grandi, Fabian Riquelme
}

\begin{document}

\maketitle    

\begin{abstract} 
{\it In this work, we seek to understand the performance of large language models in the mechanical engineering domain. We leverage the semantic data found in the ABC dataset, specifically the assembly names that designers assigned to the overall assemblies, and the individual semantic part names that were assigned to each part. After pre-processing the data we developed two unsupervised tasks to evaluate how different model architectures perform on domain-specific data: a binary sentence-pair classification task and a zero-shot classification task.  We achieved a 0.62 accuracy for the binary sentence-pair classification task with a fine-tuned model that focuses on fighting over-fitting: 1) modifying learning rates, 2) dropout values, 3) Sequence Length, and 4) adding a multi-head attention layer. Our model on the zero-shot classification task outperforms the baselines by a wide margin, and achieves a top-1 classification accuracy of 0.386. 
The results shed some light on the specific failure modes that arise when learning from language in this domain. 
}
\end{abstract}


\section{INTRODUCTION}
\label{sec:intro}
\begin{figure}[]
    \centering
    \includegraphics[width=.7\linewidth]{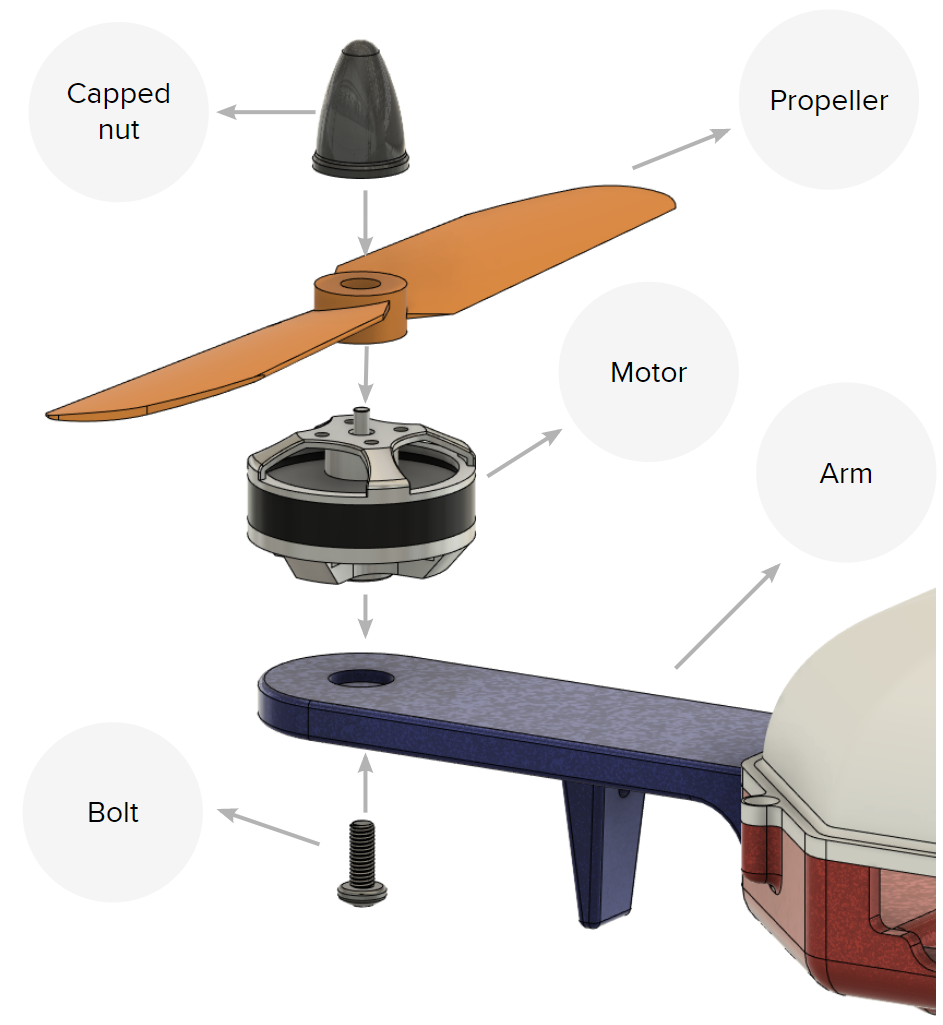}
    \caption{An example CAD assembly with each part's semantic name labeled.}
    \label{fig:assembly}
\end{figure}

In the mechanical engineering domain, natural language is used by designers and engineers throughout the design process: to express design requirements, to document design intent, and to communicate ideas and solutions to others in a complex network of people working together to create a single product. One widely used method to document and communicate design decisions is with Computer Aided Design (CAD) software. CAD software allows engineers and designers to create, modify, analyze, and optimize their design, while also documenting various aspects of the design, allowing them to communicate their design choices to others. Using CAD, designers create assemblies of parts, and often use natural language to name each individual part, as well as the assembly itself, for documentation and collaboration purposes, as shown in Figure \ref{fig:assembly}. Often, domain-specific language is used to label parts, which raises questions about the effectiveness of current natural language processing (NLP) techniques in understanding this domain-specific mechanical design language.

NLP techniques have been successfully used to understand the complexity of text and language for multiple tasks such as translation, question answering, and summarization. For these tasks, techniques such as word embeddings, transformers, and attention help navigate the unorganized structure of language. Within NLP techniques, attention-based algorithms have shown state-of-the-art results on NLP tasks as text classification outperforming most featured-based representations methods, e.g., word2vec, Glove, CoVE en ELMo \cite{yu2019improving}.
In this paper, we will explore the use of these techniques in the mechanical engineering domain. While some literature has leveraged NLP techniques for data-driven design, little work has been done specifically on natural language used to label CAD assemblies. 

To better understand the effectiveness of using Large Language Models (LLMs) in the mechanical engineering domain we will perform an NLP text classification task and recommend changes in the model to improve the results on a domain-specific corpus in the following two steps:
\begin{enumerate}
    \item We pre-process the ABC dataset to extract and clean a natural language corpus describing the assembly names and part names.
    \item We devise two unsupervised tasks to evaluate how different model architectures perform on domain-specific data: a binary sentence-pair classification task and a zero-shot classification task. 
\end{enumerate}

The code for this project can be found here\footnotemark[1].

\footnotetext[1]{\url{https://github.com/grndnl/w266_final_project}}

\section{BACKGROUND} 
\label{sec:background}
The motivation of this paper is to further explore NLP techniques in the domain of mechanical engineering and design research. The following is a review of how NLP has been applied in domain-specific applications.



\subsection{Usage of natural language data in design research}
In recent years, several datasets containing design data have been made available in the research community to support ML research in the mechanical engineering domain. These include ABC \cite{koch2019abc}, the Fusion Gallery Dataset \cite{wu20153d}, ShapeNet \cite{chang2015shapenet}, PartNet \cite{mo2019partnet}, the Mechanical Components Benchmark \cite{kim2020large}, FabWave \cite{bharadwaj2019development}, among others. These multi-modal design datasets contain 3D representations of everyday objects (furniture, vehicles, consumer electronics, industrial machinery, etc.) created by engineers and designers using CAD software. However, most of these datasets are limited by the number of classes of objects that they include, they are relatively small compared to 2D datasets such as ImageNet \cite{5206848}, and often they only contain 3D geometry information to represent individual parts (as opposed to assemblies of parts). 

Natural language is used in the mechanical engineering domain to assign semantic names to assemblies and their parts. However, of these 3D model datasets, few contain semantic information about the assemblies. One exception is the ABC dataset, which contains 1 million CAD models assembled into about 90,000 assemblies, and also provides semantic names of both the assemblies and the individual parts that make them up. The dataset was scraped from OnShape, an online CAD tool. While the 3D geometry has been used to enable the development of neural networks that learn from 3D data, such as MVCNN \cite{su15mvcnn}, PointNet++ \cite{https://doi.org/10.48550/arxiv.1706.02413}, UV-Net \cite{jayaraman2021uv}, and BREP-Net \cite{https://doi.org/10.48550/arxiv.2104.00706}, fewer work has been done with the semantic names found in 3D model datasets. 

Most research done around natural language for data-driven design focuses on sentiment analysis of consumer reviews of products to enable customer requirement prediction \cite{Feng2020}. There are some efforts to create and leverage engineering-domain-specific word embeddings: TechNet \cite{sarica2020technet}, an engineering domain-specific semantic network, was created by mining semantic relationships of elemental concepts found in US patent data and using Word2vec to create embeddings for 4 million entities. These embeddings have been leveraged for various tasks, such as data-driven design \cite{Feng2020}, function classification of parts in assemblies \cite{ferrero2022classifying}, and design idea generation \cite{sarica2021idea}.

Our work differs from prior literature by focusing on semantic information found in 3D design repositories and by leveraging the implicit knowledge of large language models.

\subsection{Applying NLP to new domains}
Bert has achieved many amazing results in NLP tasks outperforming other representation methods \cite{yu2019improving}. Prior to the existence of Transformer architectures such as Bert, there were network-based models (i.e.: recurrent, convolutional) that relied heavily on task-specific neural structures, making them highly complex and computationally expensive. Tasks related to language that require relating signals from two arbitrary inputs or outputs grow in the distance between positions \cite{vaswani2017attention} making it even more difficult to explore domain specific tasks as it will require long training on general and specific knowledge, with results that won't compare to the state-of-the-art that Bert has achieved.

Transformers like Bert have been trained on large Text Book Corpus and Wikipedia articles which give them a broader understanding of language and the contextualized word vectors allow them to understand long-distance dependencies, fine-tuning the transformer on specific data has already shown state-of-the-art performance on widely studied text classification datasets \cite{sun2019fine}. Therefore, Bert is a good starting point to be applied to new domains with little modification to its Architecture.

\subsubsection{Transformer Architecture}
 Bert follows the encoder-decoder framework using stacked multi-head self-attention and fully connected layers \cite{vaswani2017attention}. The encoder and decoder contain two sub-layers (a) a multi-head attention layer and (b) a fully connected feed-forward network, while the decoder inserts a third sub-layer which performs multi-head attention over the output of the encoder stack. In both encoder and decoder there are residual connections around each of the sub layers. This structure can be improved on specific task with a process called Bert fine-tuning.

Following previous research on Bert fine-tuning \cite{sun2019fine},  3 are the main factors at the moment of fine-tuning a Bert Model: 1) Pre-processing the data: focusing on the length of the data, token generation, sentence segmentation and language; 2) Layer Selection, each layer of Bert captures the different features of the input text, selecting the most effective layer for text classification is key, and 3) Over-fitting problem and Catastrophic Forgetting: We want to avoid the pre-trained knowledge to be erased during learning of new knowledge.

\section{METHODS}
\subsection{Data Preprocessing}
\label{sec:preprocessing}
In this work, we leverage the semantic data found in the ABC dataset \cite{koch2019abc}. Specifically, we use the \textit{assembly names} that designers assigned to the overall assemblies, and the individual semantic \textit{part names} that were assigned to each part.

To extract this useful semantic metadata from ABC, we take the raw data process of each CAD file to extract the \textit{assembly names} and \textit{part names}, yielding text data for 88,886 assemblies. We then preprocess this textual data by deduplicating any assemblies which contain exactly the name \textit{assembly name} and \textit{part names}. Informed with some EDA, we then attempt to clean the text leveraging Python RegEx operations to remove unwanted characters and common strings which were identified to not add any meaningful semantic information. We also perform this deduplication and cleaning step with the \textit{part names}, given that some part names occur very often in a single assembly. This cleaning process results in 61,725 assemblies, 48,644 of which have a unique name.

\begin{figure*}[t]
    \centering
    \includegraphics[width=\textwidth]{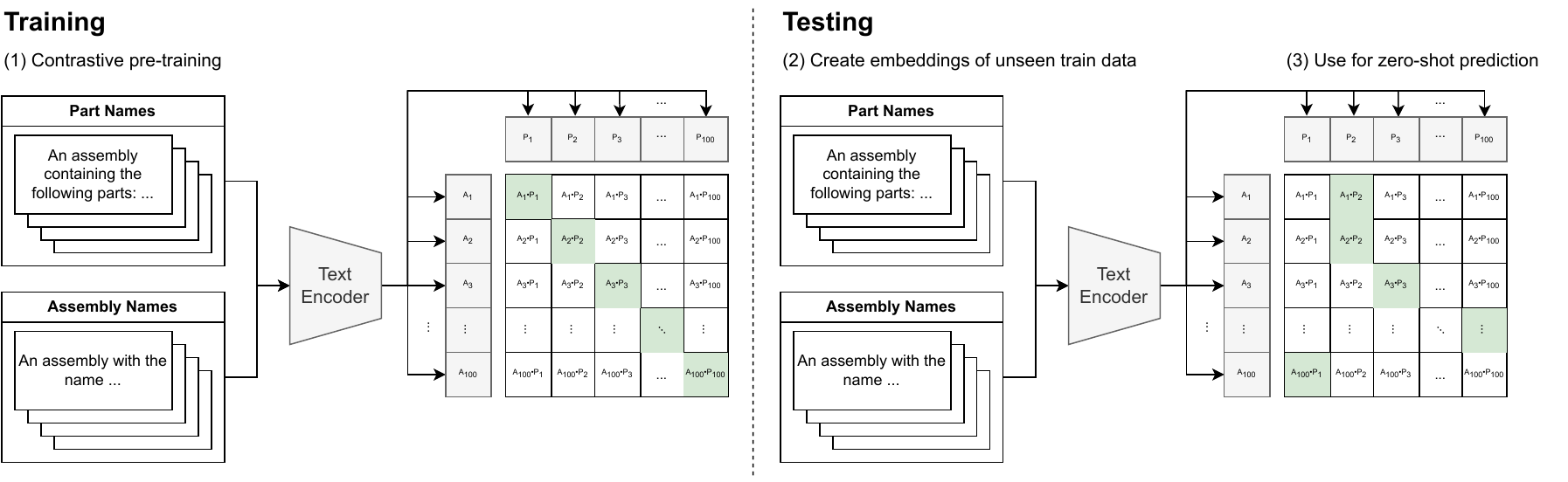}
    \caption{Summary of the assembly name prediction task. First, (1) a text encoder is contrastively pre-trained on part name and assembly name pairs. Then, (2) batches of 100 unseen part name and assembly name sentences are encoded using the same text encoder, and (3) the dot product is computed between pairs. The assembly name with the highest value is chosen as the prediction.}
    \label{fig:assembly-name-prediction}
\end{figure*}

\subsection{Understanding if elements are part of an assembly}
To mimic a practical approach to the mechanical engineering world, we are proposing a Bert-based model for a binary Text Classification, constructing auxiliary sentences to fine-tune the model with domain-specific knowledge with the help of specialized layers inspired by Bert's architecture. The prediction of this model will allow us to verify if elements are part of an assembly. The construction of the sentence is considered part of the pre-processing of the data and as mentioned before, this is a critical element to fine-tune a model \cite{michel2019sixteen}.

\subsubsection{Model Architecture}
\label{sec:model-creation}
Following the Bert architecture the model creation has the following parts:
\begin{enumerate}
    \item Input Layer: This involves the construction of the sentences for Bert fine Tuning.
    \item Bert Encoder: Pre-trained transformer. 
    \item Output Layer: Attention Multi head layer with a Binary Classification on a Dense Layer on top.
\end{enumerate}

Specifically, given an input text conforming to 2 sentences $X1$ and $X2$ , we want the classification model $f(x)= y$  to provide a result $y={0,1}$ where 0 is `not related' and 1 is `related'.

\paragraph{Input Layer}

In this study we have explored different situations on a base case LLM to understand the basic capabilities of the proposed models.
We explored 5 different cases of auxiliary sentence creation, as shown in Table \ref{tab:sentence-creation} in Appendix \ref{app:task1}, following general structure: ``An assembly named `Name of the Assembly', containing the following parts: “Parts of the assembly separated by comma”. 

The 5 different cases are tested on a Bert-based model with a binary classification on top.

\paragraph{Base Model and Fine Tuning}
As a base model we will use a Bert transformer to run a classification task and then use the following strategies to improve the results based on the literature review:

a) Learning Rate: Catastrophic forgetting \cite{sun2019fine} is a common problem in transfer learning,  values smaller or equal to 1e-4 have shown good results. For these experiments we will try learning rate values: 1e-2, 1e-3, and 1e-4.

b) Maximum sequence length: We are dealing with small-length sentences so the limitations of Bert model 512 tokens is not a problem. Reducing the size of the sentences could not only increase the testing speed but also improve the accuracy by reducing noise. Most fine-tuning studies focus on sentences containing more than 512 tokens \cite{yu2019improving}. In this study, we will evaluate the results with 128 tokens and 256 tokens on the base BERT model.

c) Number of multi-attention heads: As multi-head attention is a driving force behind Bert Architecture \cite{michel2019sixteen}, changing the amount of it could improve the results. Base Bert model has 8 Multi-head attentions on the transformer, adding a layer of multi-head attentions. Initial indications show the model improves with more heads, in this study we will test with 8 and 32 multi-heads on the attention layer.

d) Trainable Last Layer: Each layer of Bert captures the different features of the input text \cite{sun2019fine} have shown that fine-tuning the last layer of BERT gives the best performance for a text classification task. In this study, we will use this recommendation while fine-tuning the model.

e) Dropout probability: Initial experiments show that the amount of unique data is prone to over-fitting. El \cite{el2021bert} proposed that tuning the dropout hyper-parameter on a dropout layer leads to different model performance. In this study, we will experiment with a dropout of 0, 0.1, and 0.3.

\subsubsection{Evaluation}
\label{sec:evaluation}
For these experiments we will evaluate Training accuracy and Validation accuracy as we are most interested in the amount of correct answers on the binary classification.

\begin{table*}[t]
\centering
\caption{Bert fine tuning results for different hyper-parameters and architectures (train/test accuracy).}
\resizebox{0.85\linewidth}{!}{%
\label{tab:results-elements}
\begin{small}
\renewcommand{\arraystretch}{0.6}
\begin{tabular}{@{}|l|l|l|ll|ll|@{}}
\toprule
\multirow{2}{*}{\textbf{Models}} & \multirow{2}{*}{\textbf{Learning rates}} & \textbf{No Dropout} & \multicolumn{2}{c|}{\textbf{Dropout 0.1}} & \multicolumn{2}{c|}{\textbf{Dropout 0.3}} \\ \cmidrule(l){3-7} 
 &  & \textbf{Seq. Len=256} & \multicolumn{1}{l|}{\textbf{Seq. Len=128}} & \textbf{Seq. Len=256} & \multicolumn{1}{l|}{\textbf{Seq. Len=128}} & \textbf{Seq. Len=256} \\ \midrule
Base Bert & lr=0.001 & 0.546/0.621 & \multicolumn{1}{l|}{0.537/0.614} & 0.542/0.699 & \multicolumn{1}{l|}{0.547/0.592} & 0.526/0.505 \\ \midrule
\multirow{6}{*}{Base bert Model 8 heads} & \multirow{2}{*}{lr=0.01} & \multirow{2}{*}{0.565/0.515} & \multicolumn{1}{l|}{\multirow{2}{*}{0.529/0.519}} & \multirow{2}{*}{0.585/0.527} & \multicolumn{1}{l|}{\multirow{2}{*}{0.526/0.509}} & \multirow{2}{*}{0.544/0.605} \\
 &  &  & \multicolumn{1}{l|}{} &  & \multicolumn{1}{l|}{} &  \\ \cmidrule(l){2-7} 
 & \multirow{2}{*}{lr=0.001} & \multirow{2}{*}{0.722/0.627} & \multicolumn{1}{l|}{\multirow{2}{*}{0.693/0.603}} & \multirow{2}{*}{0.720/0.569} & \multicolumn{1}{l|}{\multirow{2}{*}{0.691/0.571}} & \multirow{2}{*}{0.724/0.543} \\
 &  &  & \multicolumn{1}{l|}{} &  & \multicolumn{1}{l|}{} &  \\ \cmidrule(l){2-7} 
 & \multirow{2}{*}{lr=0.0001} & \multirow{2}{*}{0.696/0.551} & \multicolumn{1}{l|}{\multirow{2}{*}{0.683/0.529}} & \multirow{2}{*}{0.718/0.611} & \multicolumn{1}{l|}{\multirow{2}{*}{0.687/0.594}} & \multirow{2}{*}{0.710/0.501} \\
 &  &  & \multicolumn{1}{l|}{} &  & \multicolumn{1}{l|}{} &  \\ \midrule
\multirow{6}{*}{Base bert with 32 attention heads} & \multirow{2}{*}{lr=0.01} & \multirow{2}{*}{0.555/0.516} & \multicolumn{1}{l|}{\multirow{2}{*}{0.563/0.533}} & \multirow{2}{*}{0.578/0.539} & \multicolumn{1}{l|}{\multirow{2}{*}{0.572/0.468}} & \multirow{2}{*}{0.527/0.602} \\
 &  &  & \multicolumn{1}{l|}{} &  & \multicolumn{1}{l|}{} &  \\ \cmidrule(l){2-7} 
 & \multirow{2}{*}{lr=0.001} & \multirow{2}{*}{0.720/0.560} & \multicolumn{1}{l|}{\multirow{2}{*}{0.610/0.582}} & \multirow{2}{*}{\textbf{0.723/0.618}} & \multicolumn{1}{l|}{\multirow{2}{*}{0.697/0.615}} & \multirow{2}{*}{0.723/0.602} \\
 &  &  & \multicolumn{1}{l|}{} &  & \multicolumn{1}{l|}{} &  \\ \cmidrule(l){2-7} 
 & \multirow{2}{*}{lr=0.0001} & \multirow{2}{*}{0.722/0.598} & \multicolumn{1}{l|}{\multirow{2}{*}{0.680/0.591}} & \multirow{2}{*}{0.710/0.531} & \multicolumn{1}{l|}{\multirow{2}{*}{0.683/0.520}} & \multirow{2}{*}{0.712/0.520} \\
 &  &  & \multicolumn{1}{l|}{} &  & \multicolumn{1}{l|}{} &  \\ \bottomrule
 \end{tabular}
\end{small}
}
\end{table*}

\subsection{Assembly name prediction}
The second task we formulate is an assembly name classification task. As Section \ref{sec:preprocessing} describes, after preprocessing there still remain 48,644 unique assembly names in our corpus. To simplify the multi-class classification task to a more reasonable 100 classes, we evaluate our model by using the text encoder as a zero-shot linear classifier to predict the assembly name of each set of 100 part names per batch, borrowing the idea from CLIP \cite{https://doi.org/10.48550/arxiv.2103.00020}. 

\subsubsection{Model architecture}
As shown in Figure \ref{fig:assembly-name-prediction}, we train the model leveraging a contrastive approach (1) where we use the same text encoder to generate embeddings for the \textit{assembly names} as well as the \textit{part names}. The network is trained to create embeddings for both sets of names, such that each \textit{assembly name} and its corresponding \textit{part name} embeddings are located near each other. Then, (2) during testing, we batch the test data into batches of 100, and generate embeddings for each \textit{assembly name} and \textit{part names} string. The order of the \textit{part names} embeddings is then randomly shuffled, and (3) we treat each batch as a 100-class zero-shot classification task by normalizing the embeddings, computing the dot product between the \textit{assembly name} and \textit{part names} embeddings, and selecting as the prediction the pairing with the highest value. We repeat the last step for each batch in the test data, and average the results.

Following the fine-tuning strategies outlined in Section \ref{sec:model-creation}, we perform a hyperparameter search by varying the number of frozen Bert layers, dropout amount, learning rate, temperature value (a parameter of the contrastive loss calculation), and Bert output token. We also try different text encoder architectures by adding additional attention layers, or additional fully-connected layers on top of Bert.

\subsubsection{Evaluation}
For this task, we chose accuracy as the evaluation metric, as each batch of 100 test classes differs from each other, it is not meaningful to average the precision and recall results of each batch. 

We evaluate our model against two baselines: a pretrained Bert base uncased model, and a Bert base uncased model that we fine-tune on the training data on a masked language model (MLM) task. 

We take these models and use them as text encoders in the testing pipeline shown in Figure \ref{fig:assembly-name-prediction} to create embeddings of the test data and compute the zero-shot prediction accuracy.

\section{Results and Discussion}
\label{sec:results}

\subsection{Entailment task}
In this section, we discuss the results of the Binary classification of auxiliary sentences of mechanical element part.

\subsubsection{Quantitative }
With the help of the auxiliary sentence creation, we evaluate the correct prediction of 
the second sentence to be the continuation of the first one. Incorrect pairings were created at random with the data from the same data set. The results of this experiment can be seen in Table~\ref{tab:results-elements}.

The experiment reveals the following:
\textbf{Learning rate}: We see an increase of both Validation Accuracy and Test accuracy as Learning Rates increase. The best results were obtained with 0.001 learning rate which was not the smallest and differs from Sun's et. al. recommendation \cite{sun2019fine}.
\textbf{Dropout}: The main tendency we observe is the increase the test accuracy but there was not significant change with the train accuracy.
\textbf{Sequence Length}: Sequence length of 256 have better results than 128. This is expected as with 128 we are removing significant information of the sentences that can be used by the model to understand relations.

\subsubsection{Qualitative }
Our evaluation shows two limiting factors: the scarce amount of data, and the large number of unique values on the dataset.
Overfitting was a common problem during training.  We observed that the results of learning rates differ from the recommendations of the literature are we are dealing with a more unique dataset, overfitting the model is easier with more epochs, and learning rates need to increase along with dropout to contrast this effect. 
Finally, an attention layer with multiple heads is beneficial to the model. This is understandable as each head will focus on a different aspect of the model at the same time, each time it will evaluate the importance of the model reducing the risk of over-fitting and at the same time encounter long-distance relations on the sentences which will help with the uniqueness of the data-set.

\begin{figure*}[]
    \centering
    \subfigure[]{\includegraphics[width=0.3\textwidth]{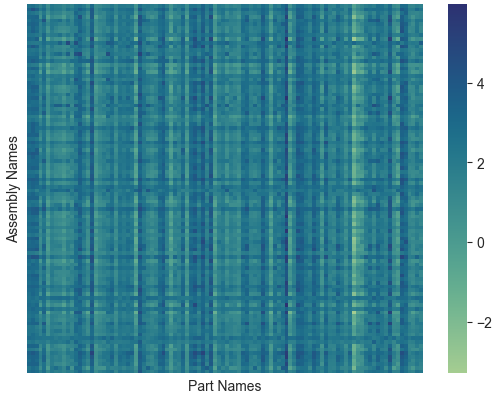}\label{fig:confusion1}} 
    \subfigure[]{\includegraphics[width=0.3\textwidth]{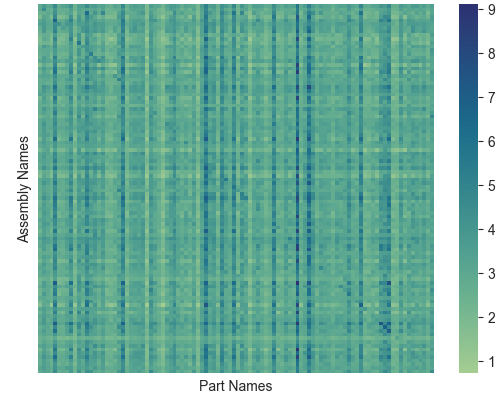}\label{fig:confusion2}} 
    \subfigure[]{\includegraphics[width=0.3\textwidth]{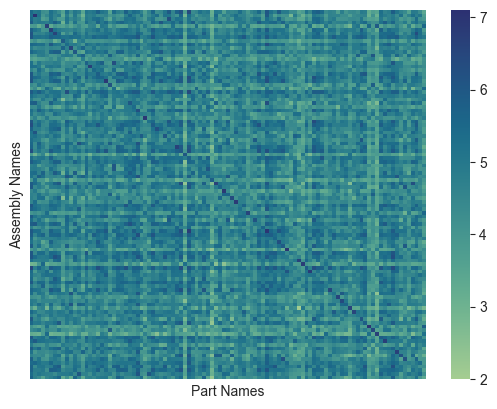}\label{fig:confusion3}}
    \caption{Cosine similarity between the normalized embeddings of the first batch of 100 assembly names vs. 100 part names for (a) Bert base, (b) Bert fine-tuned, and (c) Bert pretrained. A 100\% correct result would yield a figure where the items on the diagonal have the highest cosine similarity, indicating that the embedding of the assembly name and the embedding of the part names are most similar.}
    \label{fig:confusion}
\end{figure*}

\subsection{Assembly Name Prediction Task}
In this section, we discuss the results of the assembly name prediction task.

\subsubsection{Quantitative}
We formulate the assembly name prediction task as a 100-class classification task. To evaluate the performance of different text encoders, we calculate the mean accuracy on the 12,321 unseen examples in the test data by taking the average accuracy across 123 batches of 100 examples.

\begin{table}[h!]
\centering
\renewcommand{\arraystretch}{1}
\caption{The results of the assembly name prediction task.
}
\label{tab:assembly-name-results}
\begin{small}
\begin{tabular}{@{}cccc@{}}
\toprule
\textbf{Method} & \multicolumn{3}{c}{\textbf{Mean Test Accuracy}} \\ \midrule
\multicolumn{1}{l}{} & Top-1 & Top-5 & Top-10 \\ \cmidrule(l){2-4} 
Random Chance & 0.01 & 0.05 & 0.10 \\
Bert & 0.058 & 0.158 & 0.240 \\
Bert + Fine-tuning & 0.095 & 0.210 & 0.303 \\
Bert + Constrastive Pretraining & \textbf{0.386} & \textbf{0.553} & \textbf{0.632} \\ \bottomrule
\end{tabular}
\end{small}
\end{table}

Table~\ref{tab:assembly-name-results} shows the top-1, top-5, and top-10 mean test accuracies of the three text encoders: the Bert base uncased model, the fine-tuned Bert Base uncased model, and our pretrained Bert model using a contrastive approach. Our model outperforms the baselines by a wide margin, and achieves a top-1 classification accuracy of 0.386.

\subsubsection{Qualitative}
The results can be visually inspected by plotting the cosine similarity between the assembly name and the part name embeddings generated by each text encoder. Figure~\ref{fig:confusion} shows a batch of 100 assembly names by 100 part names in the test set. Our model shows a very distinct pattern compared to the baselines. While the baselines (a and b) showcase an interesting pattern where for each part name embedding, all assembly name embeddings are either high or low (seen as vertical lines of either high or low values), our model (c) does not showcase this pattern. In Figure~\ref{fig:confusion3}, our model can be seen to only assign a high cosine similarity value to a few assembly names for each part name embedding, which, in line with the qualitative results, seems to result in a higher overall accuracy. Also to note, is the different scales of each subplot: the better-performing models create more distance between correct and incorrect results.

To further evaluate our results, we can look at some predictions done by our model, as shown in Table \ref{tab:predictions} in Appendix \ref{app:results-assembly-name}. From the list of correct predictions, we can see that the model is very good at picking the correct assembly name when the number of parts is low (one or two). The task becomes very simple when the part name is exactly or in part the same as the assembly name; we expect the baselines to also pick up on this. For example, an assembly that contains the parts 'lens, adapter' is correctly classified with the name 'macrolens adapter for nexus 5x'.

We can also categorize the failure modes as follows. Some assemblies do not contain natural language, but rather part IDs, which we assume the model has never seen during pretraining. Some assembly names do not contain meaningful semantic information, such as `untitled document' or `liams project', which makes it very hard for the model to infer any connection with the list of part names. Different sets of assembly names could contain similarly named parts, such as `bed frame' (ground truth) and `picnic table' (predicted). Overall, the qualitative evaluation showcases how challenging this task might be even for a human, and a human baseline study would likely shed more light on the failure modes.

\subsection{Limitations and Future Work}

The fine-tuned model developed in this study could be used for the Assembly Name prediction task as an encoder. The limited data available could be augmented to get more training and testing sets.
In the future, we aim to create generative models trained on this type of data that are used to create a list of parts given an assembly name and a description of the function of the assembly, as a starting point for the mechanical designer.

\section{Conclusion}
Natural language is used by mechanical engineers to assign semantic names to both assemblies and their constituent parts during the design process, for the purposes of documentation and collaboration. This work explored the effectiveness of using large language models to understand the jargon used by designers in the mechanical engineering domain. 

To this end, we devised two unsupervised tasks: a binary sentence-pair classification task and a zero-shot classification task, which allowed us to evaluate various architectures for capturing domain-specific language. We achieved a 0.62 accuracy for the binary sentence-pair classification task, and 0.39 top-1 accuracy for the 100-class zero-shot assembly name classification task. The results shed some light on the specific failure modes that arise when learning from language in this domain. Future work could leverage the tasks we devised to assess both the quality of other design corpora or search for better model architectures to understand language in the mechanical engineering domain.

\bibliographystyle{asmems4}
\bibliography{asme2e}

\appendix       
\begin{appendices}
\onecolumn
\section{Appendices}

\subsection{Task 1: Sentence Formation}
\label{app:task1}

\begin{table}[hbt!]
\centering
\caption{}
\label{tab:sentence-creation}
\begin{tabular}{@{}llll@{}}
\toprule
\textbf{Identification} & \textbf{Sentence 1} & \textbf{Sentence 2} & \textbf{Accuracy}  \\ \midrule
Base Case & Name of  Assembly & Name of Parts & 0.469 \\
Case1 & Name of Assembly + Part1 & Name of Part 2 to the end & 0.549 \\
Case2 & Name of Assembly + Parts minus last one & Name of the last part & 0.557 \\
Case 3 & Name of Assembly + half name of parts & Name of the second half &  0.630 \\
Case 4 & Name of Assembly + Description & Name of parts & 0.594 \\ \bottomrule
\end{tabular}
\end{table}

\newpage
\subsection{Task 2: Assembly Name Predictions}
\label{app:results-assembly-name}

\begin{table}
\label{tab:predictions}
\centering
\begin{small}
\caption{A list of correct predictions and incorrect predictions done by our model.}
\resizebox{0.85\linewidth}{!}{%
\begin{tabular}{>{\hspace{0pt}}m{0.306\linewidth}>{\hspace{0pt}}m{0.102\linewidth}|>{\hspace{0pt}}m{0.321\linewidth}>{\hspace{0pt}}m{0.096\linewidth}>{\hspace{0pt}}m{0.112\linewidth}} 
\toprule
\multicolumn{2}{>{\centering\hspace{0pt}}m{0.408\linewidth}|}{\textbf{Correct Results}} & \multicolumn{3}{>{\centering\arraybackslash\hspace{0pt}}m{0.529\linewidth}}{\textbf{Incorrect Results}} \\ 
\midrule
\multicolumn{1}{>{\centering\hspace{0pt}}m{0.306\linewidth}}{Part Names} & \multicolumn{1}{>{\centering\hspace{0pt}}m{0.102\linewidth}|}{Assembly Name Ground Truth} & \multicolumn{1}{>{\centering\hspace{0pt}}m{0.321\linewidth}}{Part Names} & \multicolumn{1}{>{\centering\hspace{0pt}}m{0.096\linewidth}}{Assembly Name Ground Truth} & \multicolumn{1}{>{\centering\arraybackslash\hspace{0pt}}m{0.112\linewidth}}{Assembly Name Prediction} \\ 
\midrule
20mm stack & 20mm stack & end 1, long divider, end 2, side 1, short divider, side 2 & box shell version & ufo \\
2238 375 & 2238 375.step & mainbox, boxlid, internalcomb, basebottomkeyotherhalf, windholder, barform, basebottomkeyhalf, basetopkey & ww sport base bar box & bed frame \\
cap, sleeve & sleeve & 171, motor shaft, vt3m4x12, rail cross part, bed rotor mount, 168, 151, base board, 170, 163, 176, build plate prt11, 182, z scr & circle builder & 300 x 20 stainless chuck \\
top line holder, bottom line holder, rod carrying tube, rod shaft & line holder & sproket, spring link, bearing, stopper, shaft, spring arm, arm support & untitled document & box shell version \\
8 andr\textbackslash{}\textbackslash{}x2\textbackslash{}\textbackslash{}00e9\textbackslash{}\textbackslash{}x0\textbackslash{}\textbackslash{} mouri\textbackslash{}\textbackslash{}x2\textbackslash{}\textbackslash{}00f1\textbackslash{}\textbackslash{}x0\textbackslash{}\textbackslash{}o fig11 78 & 11.78 & motherboard panel, lower front plate, motherboard dock panel, punched angle bracket, hex bolt 3 8 1l 93190a624, hex bolt 3 8 1 2 & crop & drip disk 2 \\
tapa limaton opuesto 2, ubicaci\textbackslash{}\textbackslash{}x2\textbackslash{}\textbackslash{}00f3\textbackslash{}\textbackslash{}x0\textbackslash{}\textbackslash{}n villa luz google maps y levantamiento topografico 2, perfile limaton 2, correa & ubicación guayacanes hermanos & bras sup acier1piece & ierlgihrlihjcehveobv.step & schlüsselbrett \\
rod glue jig, v slot addon, big joint jig, rod cut jig & rod jigs & 4 laakerinen uusi malli mago & spinner mago 4 laakerinen & 300 x 20 stainless chuck \\
piggy bank plug & piggy bank plug & mxf, mpw, mtms, button, connector, pcb, intercom housing, mxa, front panel, keypad, standoff short, standoff long, spacer, 13 pi & intercom keypad controller & cap smd 2220.step \\
roller4 1 & roller 1 & index mold bottom, app index, ring mold bottom, ring mold top, ring, thumb proto base, index mold top. & finger & various 3d prints \\
lamphoudertje & houder bedlampje ikea & main support, support, ns inside panel, ns front panel, end, ns back panel, ns outside panel, side, ns bottom panel & bed frame & picnic table \\
heton ep & het & user library smd capacitor & cap smd 2220.step & pc ie mini card connector 1.5mm gap \\
804 9303 & 804 9303 & clamp, brake, grip, handle, wheel, fork & liams project & ufo \\
13mm nut, first part spanner & first part spanner & rubber foot, foot stud, foot nut & co mac 3894 swivel foot & roller 1 \\
lens, adapter & macrolens adapter for nexus 5x & case, inside & cnc bit holder & line holder \\
\bottomrule

\end{tabular}
}
\end{small}
\end{table}

\end{appendices}

\end{document}